# Cross-Task Multi-Branch Vision Transformer for Facial Expression and Mask Wearing Classification

ZHU, Armando [1*] LI, Keqin [2] WU, Tong [3] ZHAO, Peng [4] HONG, Bo [5]

[1] Carnegie Mellon University, USA
[2] AMA University, Philippines
[3] University of Washington, USA
[4] Microsoft, China
[5] Northern Arizona University, USA

*ZHU, Armando is the corresponding author, E-mail: armandoz@alumni.cmu.edu*

**Abstract:** With wearing masks becoming a new cultural norm, facial expression recognition (FER) while taking masks into account has become a significant challenge. In this paper, we propose a unified multi-branch vision transformer for facial expression recognition and mask wearing classification tasks. Our approach extracts shared features for both tasks using a dual-branch architecture that obtains multi-scale feature representations. Furthermore, we propose a cross-task fusion phase that processes tokens for each task with separate branches, while exchanging information using a cross attention module. Our proposed framework reduces the overall complexity compared with using separate networks for both tasks by the simple yet effective cross-task fusion phase. Extensive experiments demonstrate that our proposed model performs better than or on par with different state-of-the-art methods on both facial expression recognition and facial mask wearing classification task.

## 1 Introduction

Understanding human through analyzing face has been an increasingly important task for technologies interacting with humans, which allows for various application scenarios, such as robot human interaction [12, 23, 53, 55], security surveillance, and so on. It also serves as a basis for various vision tasks [7, 19, 20, 33, 43, 44]. Facial expression recognition (FER), in particular, has been extensively studied and reasonable accuracy has been achieved. However, with wearing facial masks becoming a cultural norm in the post COVID-19 era, new challenges arise where wearing facial masks impairs the ability of existing methods for facial expression recognition [14, 21]. Since facial masks occlude a big part of the human face, previous methods for facial expression recognition that assume no or minimal occlusion can be seriously affected. Therefore, the challenge to tackle face-mask aware facial expression recognition has begun to draw attention.

In this paper, we propose a model based on ViT, that allows for multi-task classification for facial expression and mask wearing classification within a unified framework. This is based on the observation of the inherent correlation between both tasks. In this case, the number of parameters required for the model is much less than the approach to classify expression and mask wearing condition with two individual networks. Moreover, to explicitly utilize information targeting both tasks, we draw inspiration from CrossViT [5] and propose a cross-task fusion module. The experiment result shows that our proposed approach outperforms existing methods while maintaining a low

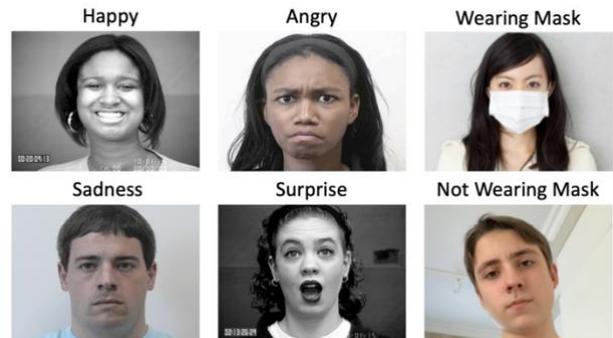

**Figure 1** Sample Images from Facial Expression Dataset, CK+, and Mask Classification Dataset, MMD

computational cost.

## 2 Related work

### 2.1 Facial Expression Recognition

Facial Expression Recognition (FER) started from handcrafted [6] solutions for handcrafted features and attributes to solutions based on deep learning [8, 9]. Turan et al. [6] proposed a system using region-based handcrafted features for facial expression recognition. More recently, since more sophisticated datasets have become publicly available, many CNN deep learning models [4, 16, 17, 32] have been proposed for this task.

Hybrid solutions have also been proposed which combines deep learning techniques with handcrafted techniques [11, 31, 35, 54]. Levi et al. [35] proposed to apply CNN on the image, with LBP features using Multi

Dimensional Scaling (MDS). More recently, many Transformer models have been introduced for different computer vision tasks. Ma et al. [11] proposed a convolutional vision transformer model that extracts features from the input image as well as form its LBP using a ResNet18.

Other researchers have thought about handling occlusion in face recognition. These approaches have been presented by [13] and [48]. An existing study on FER that takes face masks into account and recognizes emotions only from the the eyes region. This approach was evaluated on a masked FER-2013 dataset which included seven emotions [48] However, the other facial areas such as the forehead are discarded while only the eyes region is isolated using landmark detection methods, which decreases the accuracy of the FER system.

## 2.2 Mask Wearing Classification

Various types of deep learning methods are capable of the task of human mask detection [10, 15]. Jagadeeswari et al. [15] conducted an extensive review of the typical convolution neural network with different optimizers, and found out MobileNetV2 combined with Adam optimizer achieved the highest accuracy. Different approaches targeting multi-task learning have also been presented [27, 36, 37]. Ge et al. [18] propose a CNN based method, which combines two pre-trained CNNs for extracting facial regions, then described by LLE.

## 2.3 Vision Transformer

Dosovitskiy et al. [1] proposed ViT for the task of image classification. The ViT model achieves state-of-the-art performance on ImageNet for image classification, using ImageNet and JFT-300M [3] for training. Although the performance of ViT network is promising, its basic version requires a larger amount of data and has non trivial computation cost. To improve ViT models from different aspects, many variants of ViT have also been proposed, using novel tokenization process [55] or pyramid structure like CNNs [22]. T2T-ViT [55], specifically, introduces a T2T tokenization module to encode the local structural information and reduce the token size. Transformer based models [26] have also been adopted for various tasks [25, 39-41]. Cross ViT introduces a dual-path structure to extract multi-scale features to balance fine grained feature extraction with reasonable computational cost.

# 3 Methodology

In this section, we introduce the proposed solution in three separate paragraphs: some details of the ViT architecture, additive attention mechanism, and the proposed Multi-Branch Vision Transformer model.

## 3.1 Vision Transformer

The overall structure of the Vision Transformer [1]

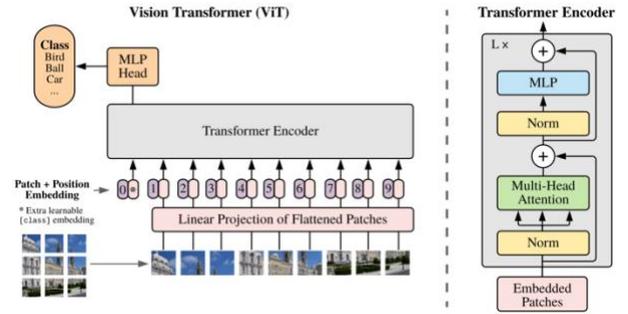

**Figure 2 Architecture of Vision Transformer**

architecture, as illustrated in Figure 2, consists of two phases: tokenization phase, and Transformer encoding phase. During the tokenization phase, image is split into $L$ fixed-size patches of size $(h * h)$ and flattened into a one-dimensional vector. Each patch s is linearly embedded into a lower-dimensional space using a learnable linear projection:

$$Embed(x_i) = x_i W_e + b_e$$

, where $W_e$ and $b_e$ are learnable weight and bias parameters. Positional encodings are added to the patch embeddings to incorporate spatial information.

Figure 2 (right) illustrates the architecture of the Transformer encoder. If consists of $L$ blocks of the attention module. Multi-head self-attention is the main part of the attention block. It is built with parallel heads of self-attention. In the architecture of a self-attention layer, keys, values, and queries all originate from the same source. This structure enables each position within the encoder to consider all positions from the encoder's prior layer. The self-attention function will be a mapping of a query ($Q$ or Q-layer) and a set of key-value ($K$ or K-layer; $V$ or V-layer) pairs to an output. The self-attention function is summarized by:

$$\text{Attention}(Q, K, V) = \text{softmax}(QK^T / \sqrt{d_k})V$$

, where $d_k$ is the dimensionality of the keys. Multi-head attention extends the self-attention by performing multiple attention operations in parallel. This is achieved by splitting the queries, keys, and values into multiple smaller vectors and applying self-attention independently to each of these smaller vectors. Afterward, the outputs of the attention heads are concatenated and linearly transformed to obtain the final output. Mathematically, multi-head attention is computed as follows:

$$\text{MultiHead}(Q, K, V) = \text{Concat}(\text{head}_1, ..., \text{head}_h)W^O$$

, where each head is computed as:

$$\text{head}_i = \text{Attention}(QW_i^Q, KW_i^K, VW_i^V)$$

, where $W_i^Q$, $W_i^K$, and $W_i^V$ are learnable weight matrices for each attention head $i$, and $W^O$ is another learnable weight matrix that combines the outputs of all attention heads.

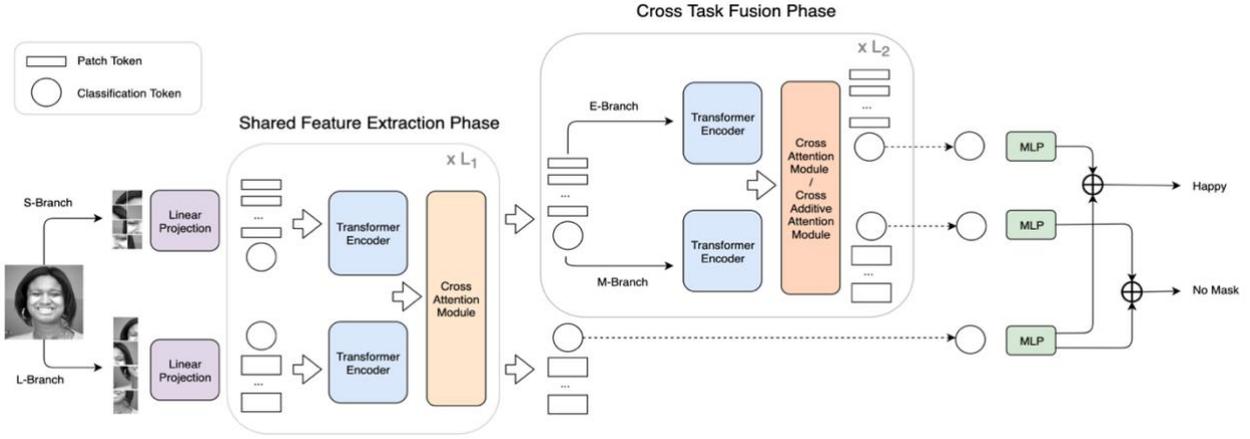

**Figure 3 Architecture of Our Proposed Pipeline**

## 3.2 Additive Attention

Additive attention, also known as Bahdanau Attention, is an alternative mechanism for computing attention scores in sequence-to-sequence models. It uses a one-hidden layer feed-forward network to calculate the attention alignment score:

$$f_{att}(h_i, s_j) = v_a^T \tanh(W_a[h_i; s_j])$$

Where $v_a$ and $W_a$ are learnable attention parameters. Here $h$ is the hidden states for the encoder, and $s$ represents the hidden states for the decoder. Additive attention provides flexibility in capturing complex relationships between query and key vectors, making it suitable for tasks where non-linear or intricate dependencies exist. The function mentioned serves as an alignment score function. In the context of a neural network, after obtaining these alignment scores, we further process them by applying a softmax function to derive the final scores.

## 3.3 Multi-Branch Vision Transformer

Figure 3 illustrates the pipeline of our proposed multi-branch vision transformer model. In brief, the model consists of two phases, unified feature extraction as the first phase, and cross task feature fusion as the second phase. This architecture allows for cross task learning while sharing the same low-level feature.

For the first phase, we adopt a dual branch ViT [5] to extract features, including two branches for different purposes: (1) L-Branch: a large branch that utilizes bigger patch size ($P_l$) with more transformer encoders, and (2) S-Branch: a small, complementary branch that operates at fine-grained patch size ($P_s$) with smaller embedding dimensions. Output of both branches are fused using cross attention module $L_1$ times, and the classification token for L-Branch is used for prediction. This dual branch architecture balances performance by incorporating fine-grained information from S-Branch, while having relatively small computational cost when processing L-Branch.

The second phase takes in outputs from S-Branch during the first phase. Two branches are introduced in this phase: (1) E-branch, branch dedicated for emotion recognition task, and (2) M-branch, branch dedicated for mask wearing classification task. Both branches share same patch size ($P_s$) as S-Branch from phase 1. The number of encoders is independently configurable for both branches. Resulting features, i.e. outputs from S-Branch are duplicated and fed to both branches. They are fused using cross attention module $L_2$ times, and the last fusion is performed using cross-additive-attention to aggregate the resulting features. Classification token from each branch is fused with classification token from L-Branch for prediction of both tasks. This design allows us to explictly exploit information and correlation between two tasks without the additional cost of having two separate networks for different tasks.

Figure 4 illustrates the cross-attention module. Specifically, for one of the two branches, it first collects the patch tokens from the other branch, and concatenates its own classification tokens. The cross-additive-attention module replaces the dot-product attention with an additive attention module to enhance the aggregation capability of the feature fusion.

We use two-stage training to finetune the whole model. For the first stage, we train the shared classifier, which is the classification token output from L-branch and S-branch, and then in the second stage, we jointly train two output classifiers from two branches and the shared classifier.

# 4 Results

## 4.1 Experimental Setup

### 4.1.1 Datasets

M-FER-2013: The FER-2013 [24] is a dataset that contains 35,887 facial images in 48x48 size in grayscale and seven different types of expressions: angry, disgust, fear, happy, neutral, sad, and surprise. We follow the automatic wearing face mask (AWFM) method used in FMA-3D [38] to form a new dataset M-FER-2013 with 35,609 images. We use the training and testing splits for training and evaluation.

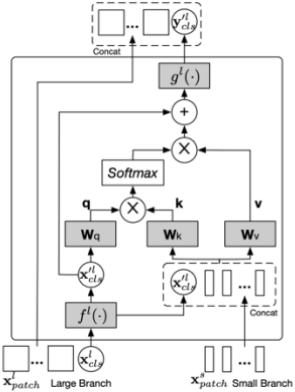

**Figure 4 Cross Attention Module**

M-RAF-DB: The Real-world Affective Faces Database (RAF-DB) [42] is published on 2017, consisting of around 30K great-diverse facial images downloaded from the Internet. We use the version with seven basic expressions for training, and use AWFM to randomly add masks with a probability of 0.5.

CK+/M-CK+: The CK+ dataset [52] consists of 593 video sequences with 10-60 frames captured from 123 subjects. In our case, we evaluate our results on the seven basic expressions to ensure a consistent comparison with different methods. We uniformly sampled and collected 1508 images for testing, and use AWFM to randomly add masks with a probability of 0.5 for M-CK+.

JAFFE: The JAFFE [46] dataset contains 213 images of different facial expressions captured from 10 different Japanese female subjects. Each subject was asked to do 7 basic facial expressions and the images were annotated with average semantic ratings. We use the dataset for validation.

MMD-FMD: [50] proposed a merged version of combining Medical Masks Dataset (MMD) and Face Mask Dataset (FMD). The MMD Dataset consists of 682 pictures with over 3k medical masked faces wearing masks, while the FMD dataset consists of 853 images. The resulting merged dataset contained 1,415 after data cleansing.

### 4.1.2 Implementation Details

In all the following experiments, we use the model architecture based on CrossViT-B [5], the base version of CrossViT with 224×224 as input imge size, 12×12 as the small patch and 16×16 for the large patch, and 3 multi-scale fusion iterations. We use the model weights pretrained on ImageNet1K [2]. All images are converted to 3 channels, resized to (224 × 224) and normalized. We use batch size of 16 and fixed learning rate ot $1 * 1.0^{-4}$, and train for 8 epochs for the first phase, and 2 epochs for the second phase.

## 4.2 Main Results

### 4.2.1 Facial Expression Recognition

In this section, we compare our proposed model trained on FER-2013 and MMD-FMD with different solutions on M-CK+ dataset, and further compare with state-of-the-art methods on JAFFE, RAF-DB and FER, which are in-the-wild datasets. It can be seen from Table 1 that accuracy on scenarios with masks is significantly affected. Under this case, our proposed model outperforms other traditional and handcrafted methods and improves the accuracy based on CrossViT. We achieve this with only a small increase in FLOPs and model parameters compared with a two-fold increase for CrossViT model (42.4 GFLOPS and 209.4M Parameters).

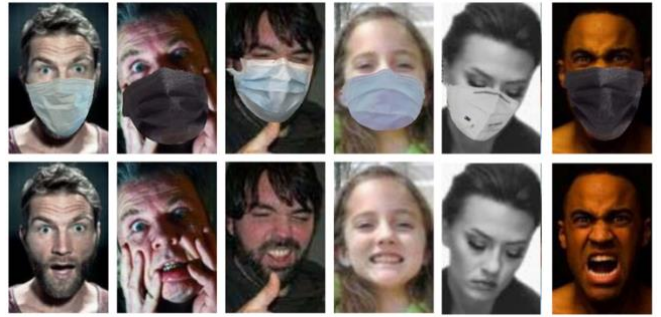

**Figure 5 Sample images from M-RAF-DB with mask (up) and without mask (bottom)**

**Table 1 Comparison with Different Methods**

| Model | Accuracy | | FLOPs | Params |
|---|---|---|---|---|
| | CK+ | M-CK+ | | |
| **Handcrafted [6]** | 0.9503 | 0.6359 | N/A | N/A |
| **VGG19 [30]** | 0.9658 | 0.6847 | 19.7G | 144M |
| **ResNet50 [45]** | 0.9723 | 0.7003 | 25.7G | **4.3M** |
| **ViT [1]** | 0.9830 | 0.744 | **17.6G** | 86.7M |
| **Cross ViT [5]** | 0.9865 | 0.7532 | 21.2G | 104.7M |
| **Proposed** | **0.9856** | **0.7702** | 24.6G | 125.8M |

Table 2 shows more quantitative results on different datasets compared with state-of-the-art methods, based on either CNN or visual transofmer. On both lab datasets and wild datasets, our model consistently achieves good results over other methods. We have the highest accuracy on M-JAFFE with a 70.59% acuraccy and 77.85% on M-FER-2013, both on scenarios where mask is randomly applied.

**Table 2 Comparison with State-of-the-Art (Accuracy)**

| Model | M-JAFFE | M-RAF-DB | M-FER-2013 |
|---|---|---|---|
| **ResNet50 [45]** | 0.6602 | 0.6126 | 0.6978 |
| **ViT [1]** | 0.688 | 0.643 | 0.7519 |
| **Cross ViT [5]** | 0.6994 | 0.6479 | 0.7632 |
| **Ours** | **0.7059** | **0.6438** | **0.7785** |

### 4.2.2 Facial Mask Detection

We compare our performance of mask classification with existing methods. The performances of different methods compared with our proposed network can be found in Table 3. Generally, CNN-based networks outperform the traditional SVM method, while Transormer based methods outperform CNN methods. Our method showed slight improvements over the ViT network while sharing the underlying network with FER task.

**Table 3 Evaluation Results on MMD-FMD (Accuracy)**

| Model | Accuracy |
|---|---|
| SVM RBF | 0.8527 |
| VGG19 [30] | 0.9459 |
| ResNet50 [45] | 0.9675 |
| ViT [1] | 0.9774 |
| Ours | 0.9793 |

### 4.3 Ablation Studies

**Different attention methods.** We compare different settings of our network, substituting the cross-addition-attention fusion method with cross-attention method which uses dot-product attention. Table 4 shows that cross-addition-attention achieves higher accuracy on both M-CK+ and M-JAFFE dataset. Intuitively, additive attention module is a better fit for the scenario of feature aggregation and final prediction.

**Importance of the cross-task fusion module.** We also experiment with predicting from fusing the output directly from both L-Branch and S-Branch from phase 1 while training for both tasks. Result on Table 4 reveals that the model degrades without using the output from the second phase where cross attention is applied, which is 2% worse than our proposed architecture. This shows the effectiveness of the cross-task learning and prediction phase.

**Table 4 Evaluation Result on Different Network Configurations**

| Model | Accuracy | | FLOPs | Params |
|---|---|---|---|---|
| | M-CK+ | M-JAFFE | | |
| Ours with dot-product attention | 0.7689 | 0.7033 | 24.6G | 125.8M |
| Ours with output from first phase | 0.7485 | 0.6859 | 21.2G | 104.7M |
| Ours | 0.7702 | 0.7059 | 24.6G | 125.8M |

## 5 Conclusion

In this paper, we present the Cross-Task Multi-Branch Vision Transformer, a novel multi-branch vision transformer for learning mult-tasks features, to improve the accuracy of facial expression recognition and mask detection tasks. We utilize the similarity of both classification tasks that are able to share a joint representation, and then by using the proposed cross task fusion stage, task specific representation can be obtained while exchanging information by cross attention modules. By conducting extensive experiments, we demonstrate our proposed model performs better than or on par with several recent works, in addition to other baseline methods. In future work, we aim to generalize the network architecture to provide a solution for more generalized and competitive tasks.